\documentclass[conference]{IEEEtran}
\IEEEoverridecommandlockouts
\usepackage{times}
\usepackage{graphicx}
\usepackage{latexsym}
\usepackage{cite}
\usepackage{amsmath,amssymb,amsfonts}
\usepackage[ruled,vlined]{algorithm2e}
\usepackage{algorithmic}
\usepackage{graphicx}
\usepackage{textcomp}
\usepackage{float}
\usepackage[caption=false]{subfig}
\usepackage{multirow}
\usepackage{xcolor}
\usepackage{authblk}
\usepackage{hyperref}

\newcommand{\R}{\mathbb{R}}

\def\BibTeX{{\rm B\kern-.05em{\sc i\kern-.025em b}\kern-.08em
    T\kern-.1667em\lower.7ex\hbox{E}\kern-.125emX}}
\begin{document}
	
\title{Adversarial Signal Denoising with Encoder-Decoder Networks}

\author[1]{Leslie Casas\thanks{Leslie Casas thanks the Peruvian Innovation Program for Competitiveness and Productivity - Innovate Peru (grant BECA-1-P-240-13) for supporting her doctoral studies. Project page: \url{http://campar.in.tum.de/Chair/ProjectAdversarialDenoising}}}
\author[2]{Attila Klimmek}
\author[1]{Nassir Navab}
\author[2]{Vasileios Belagiannis}
\affil[1]{Technische Universit\"at M\"unchen, Garching bei M\"unchen, Germany}
\affil[2]{Universit\"at Ulm, Ulm, Germany}
\affil[ ]{\normalsize \textit{firstname.lastname@\{tum.de, uni-ulm.de\}}}

\maketitle
\thispagestyle{plain}
\pagestyle{plain}

\begin{abstract}
The presence of noise is common in signal processing regardless the signal type. Deep neural networks have shown good performance in noise removal, especially on the image domain. In this work, we consider deep neural networks as a denoising tool where our focus is on one dimensional signals. We introduce an encoder-decoder architecture to denoise signals, represented by a sequence of measurements. Instead of relying only on the standard reconstruction error to train the encoder-decoder network, we treat the task of denoising as distribution alignment between the clean and noisy signals. Then, we propose an adversarial learning formulation where the goal is to align the clean and noisy signal latent representation given that both signals pass through the encoder. In our approach, the discriminator has the role of detecting whether the latent representation comes from clean or noisy signals. We evaluate on electrocardiogram and motion signal denoising; and show better performance than learning-based and non-learning approaches.
\end{abstract}


\section{Introduction}
In signal processing, the presence of noise is a common problem regardless the signal type. One way to recover the signal is to use neural networks~\cite{arsene2019deep, im2017denoising}. This approach has been particularly popular in the image domain where learning-based approaches (e.g. denoising autoencoders~\cite{2010_vincent} or encoder-decoder networks~\cite{agostinelli2013adaptive, 2016_mao}) have advanced the field. Similarly in audio and speech processing, the recent advances of deep neural networks have resulted in promising results~\cite{lu2013speech, 2016_vandenoord, grill2017two, rethage2018wavenet}. Contrarily, the influence of learning-based methods is rather limited on lower dimensional signals such as motion.

For lower dimensional signals, the prior work mainly consists of non-learning methods such as filtering, wavelet transforms and empirical mode decomposition. Linear denoising methods that rely on filtering, e.g.~Wiener filter~\cite{antoniou2016digital}, work well in the presence of stationary noise, but they have shown limitations when the signal and noise share the same spectrum~\cite{boudraa2004emd}. In wavelet transforms, the performance depends on the choice of the predefined basis functions, which may not reflect the signal's nature~\cite{kopsinis2009development}. Finally, empirical mode decomposition~\cite{huang1998empirical} is a data-driven approach that works with stationary and non-stationary signals. However, it can face difficulties in decomposing the signal into unique frequency components, resulting in mode mixing~\cite{huang2014hilbert}. 

\begin{figure}[t]
	\centering
	
	\subfloat[Beginning of training (motion).]{
		\includegraphics[width=0.24\textwidth]{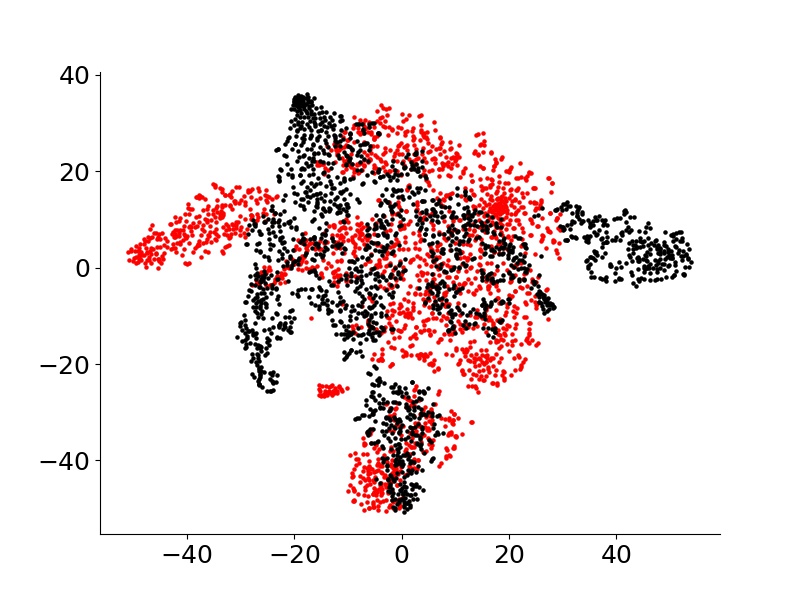} 
		\label{p_D_req_100}
	}
	\subfloat[End of training (motion).]{
		\includegraphics[width=0.24\textwidth]{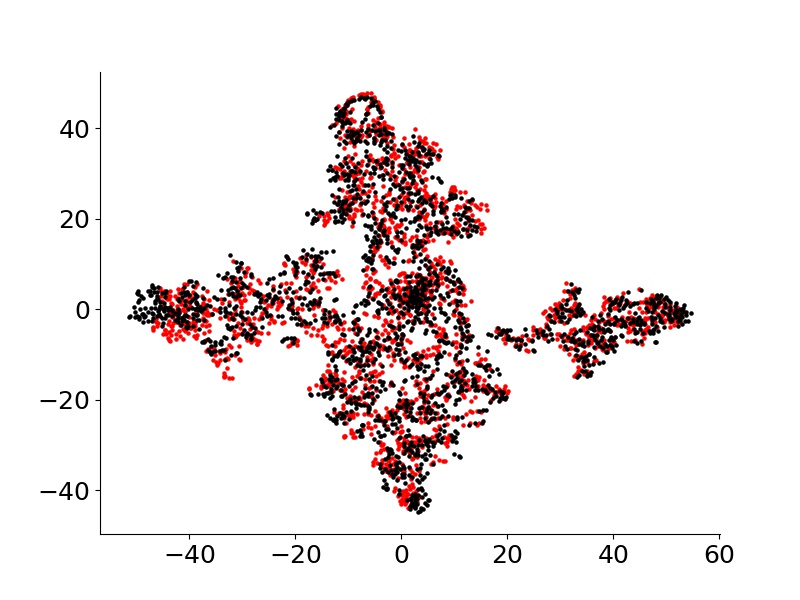}
		\label{p_D_req_400}
	} \\
	\subfloat[Beginning of training (ECG).]{
		\includegraphics[width=0.24\textwidth]{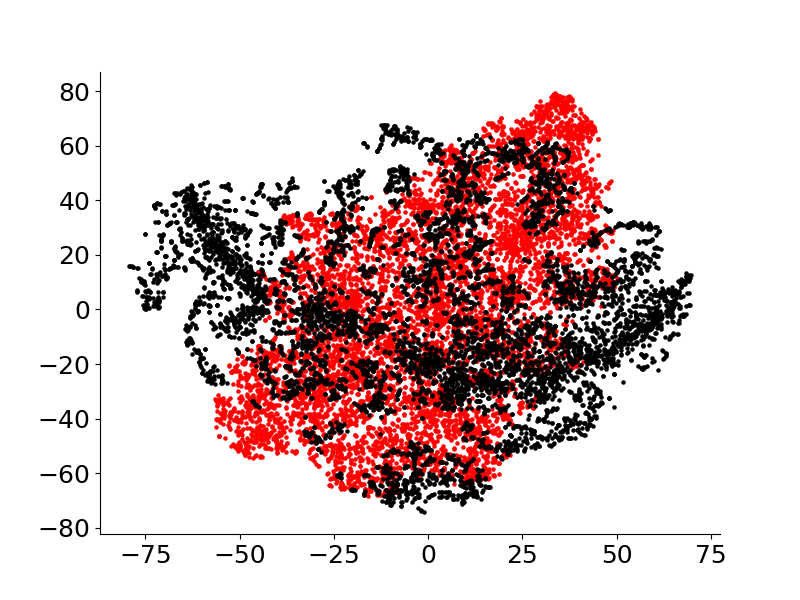} 
		\label{p_D_req_100_2}
	}
	\subfloat[End of training (ECG).]{
		\includegraphics[width=0.24\textwidth]{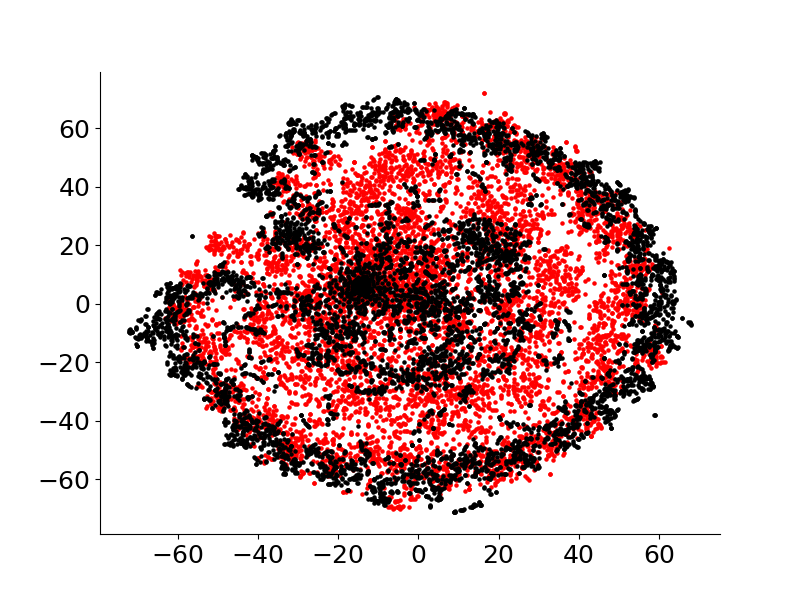}
		\label{p_D_req_400_2}
	}
	\caption{Signal Distribution Alignment. We visualize the clean (red) and noisy (black) signal latent representations for the motion and electrocardiogram (ECG) over time. We make use of $t$-SNE~\cite{maaten2008visualizing} to project the latent representation to a two dimensional space. Although we visualize the same clean samples, they are distributed differently because the encoder parameters are keep changing over training time.
	}
	\label{fig:p_graph}
\end{figure}

In this work, we follow the data-driven paradigm too. However, we build on deep neural networks and adversarial learning. We treat signal denoising as a distribution alignment task. Then, we present an adversarial encoder-decoder network architecture for denoising signals that are represented by a sequence of measurements. In our approach, a discriminator network classifies the signal into noisy or clean, given the signal's latent representation input. Aligning the clean and noisy signal distributions is equivalent to removing the noise. Unlike the standard GAN training~\cite{goodfellow2014generative} and adversarial autoencoders~\cite{makhzani2015adversarial, creswell2018denoising}, we propose a different formulation that suits-well to our problem. We propose to pass the clean signals through the encoder. Afterwards, we use the latent representation of the clean and noisy signals as input to the discriminator. Our model learns to align the noisy signal latent distribution with the respective clean signal distribution that eventually acts as denoising.

Our motivation for the design of the encoder-decoder network comes from the advances in the image domain. First, we adopt the structure of a fully convolutional network (FCN)~\cite{long2015fully} for one-dimensional data. We design the encoder to denoise the input and transform it to the latent representation. On the other hand, the decoder reconstructs the clean signal from the latent representation. To facilitate the reconstruction, we introduce residual learning with shortcut connections from the encoder to the decoder. Moreover, we introduce dilated convolutions \cite{chen2018deeplab} for the encoder and dilated deconvolutions for the decoder in order to increase the effective receptive field of the network. This allows us to efficient process temporal data without employing recurrent models, as shown in~\cite{2018_bai}. While these operations are well-established in the image domain, they have not been sufficiently explored for one-dimensional signal processing yet.

In summary, our work makes the following contributions: (i) an adversarial encoder-decoder network for one-dimensional signal denoising, (ii) generalization to different signal and noise types and (iii) better performance than prior work.

\begin{figure}[t]
	\centering
	\includegraphics[width=0.47\textwidth]{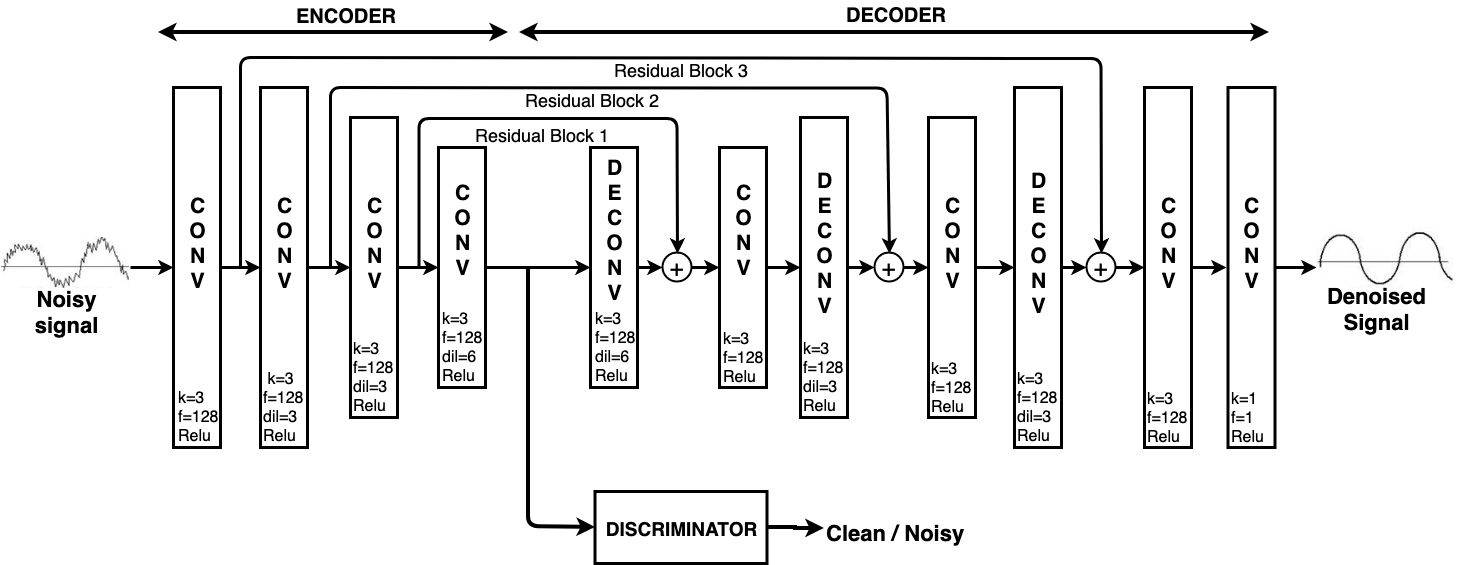}
	\caption{Approach Overview. The input to the network is noisy signal, while the output is the clean version of it. In addition, the encoder feeds the discriminator with the latent representation of the clean and noisy signals.}
	\label{AdvEncDecModels}
\end{figure}

\section{Adversarial Signal Denoising}
Let $\mathbf{x}\in \R^{D}$ be the corrupted version of the one-dimensional signal $\mathbf{y} \in \R^{D}$, where $D$ represents the signal length. Our goal is to estimate the clean signal $\mathbf{y}$ with the function $f:{\R}^{D}\rightarrow {\R}^{D}$ that is the composition of an encoder and decoder network, given by $f(\mathbf{x}) = \psi(\phi(\mathbf{x}))$. The encoder $\phi:{\R}^{D}\rightarrow {\R}^{K}$ provides a denoised latent $K$-dimensional representation of the input. The decoder $\psi:{\R}^{K}\rightarrow {\R}^{D}$ reconstructs the denoised signal from the latent space. The whole network is parametrized by $\boldsymbol{\theta}$, where the parameters are learned with back-propagation and stochastic gradient descent. Our contribution is on the parameter learning and in the loss function in particular.

First, we rely on a standard loss function for autoencoders, as it is presented in Sec.~\ref{AENet}. Both the encoder and decoder of our network architecture are trained with encoder-decoder loss. Second, we introduce adversarial learning for the encoder. To this end, a discriminator network is proposed for detecting clean and noisy signals. Our adversarial formulation is presented in Sec.~\ref{AdvLearn}.

\subsection{Encoder-Decoder Model}\label{AENet}
Given a set of training data, the encoder-decoder objective that we aim to minimize is 

\begin{equation}
	\mathcal{L}_{ED}(f)=\mathbb{E}_{\mathbf{x},\mathbf{y}}[f_{\boldsymbol{\theta}}( \mathbf{x})-\mathbf{y}]^{ 2 }.
	\label{AEloss}
\end{equation}
The encoder's and decoder's parameters $\boldsymbol{\theta}$ correspond to convolution and deconvolution operations. We model the encoder with dilated convolutions and the decoder with dilated deconvolutions (i.e.~transposed convolutions). We carefully design both parts so that they are symmetric. Then, the shortcuts connect the encoder with the decoder as illustrated by Fig.~\ref{AdvEncDecModels}. Next, we present in detail the dilated convolution, dilated deconvolution and shortcut operations for one-dimensional data.

\paragraph{Dilated Convolutions.} Dilated convolutions increase the effective receptive field of the network without increasing the number of parameters. In our model, we introduce dilated convolutions in the encoder network. Given the 1D input signal $\textbf{x} \in \R^{M}$ and a 1D filter $\mathbf{w} \in R^{r}$, the dilated convolution at the position $t$ is defined as $\textbf{y}[t] = \sum _{ j=0 }^{r-1}{ \textbf{x}[t+d\cdot (j-1)] \cdot \mathbf{w}[j] }$, where $d$ is the dilation factor that we linearly increase in the encoder, while we fix the size of the filter. Furthermore, the 1D convolution kernel is centered at $t$.

\paragraph{Dilated Deconvolutions.} Our objective is to build a symmetric decoder to the encoder which is defined by transposed operations. We propose the dilated deconvolution to upsample the latent and feature representations. Assuming the 1D input $\textbf{x}$ and a 1D filter $\mathbf{w}$ again, the dilated deconvolution at $t$ is now defined as $\textbf{y}[t] = \sum _{\substack{j=0}}^{r-1}{ \textbf{x}[t-d\cdot j] \cdot \mathbf{w}[j] }$, where $j\leq \frac{t}{d}$ and $j\geq \frac{t-\left | \textbf{x} \right |}{d}$ in order to avoid indices out of \textbf{x} range. For a 1D input, the operation is similar to inverting the filter and applying it to the signal. The dilation  factor $d$ is also symmetric. This means that the decoder starts with larger factors that linearly decreases at every upsampling.

\paragraph{Residual Blocks.} Residual learning~\cite{2016_he} has been introduced for building very deep neural networks, without vanishing gradient problems. A residual block can be represented as: $\textbf{x}_{\ell+1} = F(\textbf{x}_{\ell}) + \textbf{x}_{\ell}$, where $\textbf{x}_{\ell}$ is the input of the $\ell$-th layer and $F$ is the residual mapping. The mapping is usually a set of operations such as convolution, activation and batch normalization. These operations are followed by addition with a skip(shortcut) connection. Here, we propose residual blocks to connect the encoder with the decoder. Each block includes operations from the previous one. We introduce a combination of convolutions and deconvolutions inside the block. The residual learning contributes to reconstructing the denoised signal from the latent space.

\paragraph{Encoder-Decoder Architecture.} The input $\mathbf{x}$ is passed through the encoder and then through the decoder (Fig.~\ref{AdvEncDecModels}). The encoder is composed of a standard 1D convolution followed by three levels of dilated convolution with 3, 3 and 6 dilation factors. The decoder has a symmetric structure with a set of three levels of dilated deconvolutions with symmetric dilation factors that are 6, 3 and 3, followed by a standard 1D convolution that results in the clean signal reconstruction. The convolutions use padding to retain the data size. Furthermore, the first convolution adds 128 feature channels to the input. The same number of features propagates along the encoder-decoder network, while the last convolution decreases the number of channels to 1. Note that all convolutions and deconvolutions have kernel size 3 and are followed by a non-linearity (ReLUs). Only the last convolution has a linear activation. The shortcut connections take place after each deconvolution layer, where the features of the lower layers are added to the symmetric features of the deconvolution layers. After each addition, there is a convolutional layer that weights the contribution of the shortcut connection (see Fig.~\ref{AdvEncDecModels}). At the end, the encoder-decoder network outputs the denoised signal with the same dimensions as the noisy input. The proposed network architecture is next combined with a discriminator neural network as part of the adversarial learning.

\subsection{Adversarial Learning for Denoising}~\label{AdvLearn}
Adversarial learning has been established through Generative Adversarial Networks (GANs)~\cite{goodfellow2014generative} for image generation. The idea is to build an image generation network, using latent variables as input; and obtain supervision on the reconstructed image from another neural network: the discriminator. The discriminator's role is to distinguish generated from real images, where the overall objective is to generate images that fool the discriminator, i.e. generate images that are indistinguishable from the real ones.

In our problem, there are clean and noisy signals instead real and fake images. In addition, we experienced empirically that making use of the raw signals as input to the discriminator does not work well. We thus rely on the latent representation of the signals as discriminator input. Consequently, both clean and noisy signals have to pass through the encoder to obtain the latent representation. During parameter update, though, the gradients of the noisy signals contribute to the parameter update of the encoder network. Our objective is represented as adversarial training, but it differs from the original GANs~\cite{goodfellow2014generative} or adversarial autoencoders~\cite{makhzani2015adversarial}.

\paragraph{Objective}
We treat the task of denoising as distribution alignment, where the misalignment occurs because of signal noise. The two players are the noisy and clean signal. The encoder $\phi(\cdot)$ receives the noisy sample $\mathbf{x}$ as input to produce its latent representation. Therefore, the role of the generator is implicitly assigned to the encoder. In addition, the encoder $\phi(\cdot)$ is used for generating the latent representation of the clean signal $\mathbf{y}$. Unlike standard GAN problems, the real data distribution is not given here.

The discriminator $DS(\cdot)$ classifies the latent representation into clean or noisy. Finally, fooling the discriminator in adversarial learning means to align the latent representations of the two signals. For that reason, adversarial learning acts as denoising. The proposed model is shown in Fig.~\ref{AdvEncDecModels}. Following the original formulation from~\cite{goodfellow2014generative}, we define the objective as:
\begin{align}\label{cGANeq}
	\mathcal{L}_{GAN}(\phi,DS) &=   \mathbb{E}_{\mathbf{y}}\left[\log (DS(\phi(\mathbf{y})))\right] \nonumber \\
	&\qquad +\mathbb{E}_{\mathbf{x}}\left[\log (1-DS(\phi(\mathbf{x}))\right].
\end{align}

Note that the first term (i.e. clean data) makes use of the encoder (i.e. generator).  To avoid updating the encoder with the discriminator's gradients of the clean signal, we introduce $\lambda \in \{ 0,1\}$ as control term and reformulate the objective as:
\begin{align}\label{cGANeq2}
	\mathcal{L}_{Adv}(\phi,DS, \lambda) &=  \lambda \mathbb{E}_{\mathbf{y}}\left[\log (DS(\phi(\mathbf{y})))\right] \nonumber \\
	&\qquad +\mathbb{E}_{\mathbf{x}}\left[\log (1-DS(\phi(\mathbf{x}))\right].
\end{align}
The control term is set to zero $\lambda=0$ when updating the encoder network $\phi(\cdot)$. The adversarial objective is given by: \begin{align}
	\arg \min_{\phi,f,\lambda=0}\max_{DS, \lambda=1} \mathcal{L}_{Adv}(\phi,DS, \lambda).\label{adv_objective}
\end{align}

\subsection{Complete Objective }
The training of the encoder-decoder network is based on Eq.~\ref{AEloss} and Eq.~\ref{adv_objective}. The final objective that includes the adversarial and encoder-decoder terms is 
\begin{align}
	\arg \min_{\phi,f,\lambda=0}\max_{DS, \lambda=1} \mathcal{L}_{Adv}(\phi,DS, \lambda) + \mathcal{L}_{ED}(f).\label{full_objective}
\end{align}
The two terms could be weighted by a constant. However, we observed similar performance when balancing the two terms and we thus skip it. During the maximization, the discriminator $DS(\cdot)$ is updated using clean and noisy latent representations, while the encoder-decoder network $f(\cdot)$ is updated during the minimization. The encoder part $\phi(\cdot)$ is additionally updated with gradients from the discriminator.

\subsection{Discriminator Design}
The input to the discriminator is the latent representation given by the encoder. We have found empirically that a 4-layer discriminator is sufficient for our problem. This architecture is also similar to discriminators for compression~\cite{belagiannis2018adversarial} or domain adaptation~\cite{tzeng2017adversarial}. There is first a convolution to reduce the channel dimensions to one, followed by two fully connected layers with 150 units and ReLU activation, each. Finally, the signal is reduced to binary classification with the last fully connected layer and using sigmoid activation.

\section{Experiments}\label{sec:experiments}

Our evaluation on one-dimensional signal denoising is based on motion and electrocardiogram (ECG) signals. In both cases, we compare our results with learning-based and non-learning approaches; and provide a  model component analysis.

\noindent \textbf{Related approaches.} For the comparison with prior work, we consider  wavelets~\cite{tagkey2009iii}. We obtained the best parameters after an exhaustive search with the modified overlap wavelet transform, using Symlets 8 with 5 levels of decomposition, soft thresholding and level-dependent noise estimation. In addition, we compare with learning-based denoising approaches. We implement the denoising autoencoder (AE)~\cite{2010_vincent} with three layers for encoding and another three for decoding. Note that we trained deeper AE models, but there was not an improvement of the results. Next, an LSTM architecture~\cite{greff2017lstm} with two cells is included for comparisons with recurrent neural networks. Lastly, we build a variant of a WaveNet~\cite{2016_vandenoord}, originally used for speech denoising~\cite{rethage2018wavenet}. The evaluation metric is the signal-to-noise ratio (SNR) for all cases. The initial noise is reported as reference for the level of improvement after denoising.

\noindent \textbf{Implementation.} We choose the encoder-decoder network to have a 3-layer encoder and 3-layer decoder. We tried different layer variations, but we empirically found that the 3-layer model suits well for the examined signals. The same model is employed for all experiments. We first evaluate the encoder-decoder network based only on Eq.~\ref{AEloss} without adversarial learning, similar to standard autoencoder. Second, we evaluate our complete model with adversarial learning based on Eq.~\ref{full_objective}. Below, we discuss the results for each experiment. For all our models, we rely on the AdaDelta\cite{2012_zeiler} optimizer with weight decay 5e-4. The weights of the convolutional and deconvolutional layers are initialized with Glorot uniform distribution~\cite{2010_glorot} and hyperparameters are obtained by grid search. The network input for all models is raw data. Finally, we set the temporal window to 10 measurements during training, while the inference works with adaptive input.

\subsection{Motion Signal Evaluation}
We select the European Robotics Challenge(EuroC) MAV dataset~\cite{2016_burri} that consists of 11 sequences of inertial measurement unit sensors and motion capture data. Each sequence contains angular velocity and acceleration measured at 200Hz, while the 3D position and angular velocity are obtained with a motion capture system. The noise of this signal is composed mainly of Gaussian and random walk noise. We denoise the angular velocity, because it is the only one with available ground-truth. The velocity has 3 dimensions, which we treat as a sequence of measurements. For the evaluation, we perform an 11-fold leave-one-out cross-validation.

In Table~\ref{tab:real-eval}, the average results are summarized using the SNR metric. Our baseline is the encoder-decoder network. Our adversarial encoder-decoder network has the same parameters with the baseline at inference, but the parameters are more during training due to the discriminator network. The best result is highlighted in bold. The AE performs well in increasing the SNR from 12.57dB to 23.48dB. The performance of WaveNet is on the same level with AE, while the LSTM  performance is not as good as the learning-based models. Besides, we observe that our encoder-decoder already denoises better than the other approaches. The introduction of the adversarial learning further improves the results to 32.08 SNR. Finally, we also experienced that the wavelet method cannot adequately cope with unknown type of noise.

We further provide a visualization of the latent space of our model, projected to two dimensions with t-SNE~\cite{maaten2008visualizing}. In Fig.~\ref{fig:p_graph}, we show the projected samples at the beginning and at the end of training. The clean samples are shown in red and the denoised samples are the black dots, respectively. Although the noisy and clean data have the same range of values in the beginning, they are clearly misaligned. At the end of training, Fig.~\ref{fig:p_graph} shows the alignment between the two data distributions. Note that the projection of the clean samples differs between the beginning and  end of training since we make use of the encoder to generate their latent representation.

\begin{table}
	\centering
	\caption{Denoising results on Motion and ECG datasets.}
	\label{tab:real-eval}
	\resizebox{0.47\textwidth}{!}{
		\begin{tabular}{|l|c|c|c|}
			\hline
			&\# Param.& \multicolumn{2}{c|}{SNR(dB)}\\ \hline
			& &Motion&ECG\\ \hline
			Initial Noise&-&12.57&-6.72\\ \hline
			Wavelets&-&12.79&-5.90\\ \hline
			AE~\cite{2010_vincent}  	&	102.855& 23.48&4.67	 \\ \hline
			LSTM~\cite{greff2017lstm}    &62.155	& 19.11&2.65   \\ \hline
			WaveNet Denoiser~\cite{rethage2018wavenet} &463.747& 23.33&4.45	  \\ \hline 
			Our Encoder-Decoder (Eq.~\ref{AEloss}) &444.161	& 25.21 &4.24 \\ \hline	
			Our Adversarial& & &\\
			Encoder-Decoder (Eq.~\ref{full_objective})&468.141 &\textbf{32.08}&\textbf{5.30}\\ \hline
	\end{tabular}}
\end{table}

\subsection{Electrocardiogram (ECG) Evaluation}
Next, we explore the generalization of our approach to denoise another type of one-dimensional signal. We choose the Physionet ECG-ID database~\cite{goldberger2000physiobank} that has 310 ECG records from 90 subjects. Each record contains the raw ECG signal and the manually filtered ground-truth version. Our sampling frequency is similar to the motion signal and thus the same network architecture is suitable for the experiment. The dataset does not have a standard evaluation protocol. For that reason, we randomly choose 10 subjects for test set and rely on the rest data for training and validation. Here, this signal is often corrupted by power line interference, contact noise and motion artifacts. Denoising now becomes more complex, because the ECG signal is non-stationary and has overlapping spectrum with the noise. The results are presented in Table~\ref{tab:real-eval}.

In this evaluation, we make similar observations as with the motion signals. AE and WaveNet perform similarly well in terms of SNR performance. The LSTM model is able to denoise too, but it is again behind the other learning-based models. Finally, the non-learning algorithm, i.e.~wavelets, has difficulties with denoising this data. Although it removes some noise, the result is behind the other approaches. Finally, our encoder-decoder model delivers similar denoising results to AE and WaveNet. Our complete model (encoder - decoder + adv. learning), though, achieves much higher SNR compare to all other approaches.

\subsection{Model Component Analysis} \label{compoAnalysis}
The proposed adversarial encoder-decoder is composed of dilated convolutions and deconvolutions, residual connections and the adversarial objective. Here, we analyze the contribution of each component to the final model. We define a base model for evaluation and then gradually add the proposed components. For the evaluation, we consider again the Physionet database~\cite{goldberger2000physiobank} and the angular velocity from~\cite{2016_burri}.

In Table~\ref{compo-eval}, we present the results incrementally adding the model components. The best result is highlighted in bold. The base model is composed of a 2-layer encoder-decoder network (4 layers in total). It has standard convolutions, without dilation. Then, an extra layer is added to the encoder and another to the decoder (6 layers in total). Next, the dilated convolutions and dilated deconvolutions are introduced, followed by the residual learning. At the end, the adversarial learning is included in the model training. We also tried a 4-layer model that did not give further improvements for the examined signals. The largest noise reduction occurs when introducing the dilated convolutions and dilated deconvolutions in our model, from 22.60dB to 24.47dB on the angular velocity and from 3.49dB to 5.09dB on the ECG data. We also observe similar increase of the SNR when introducing the adversarial learning. Note that we trained our model without residual connection due to the performance drop in the ECG evaluation. However, it results on overall better performance to make use of the residual connections.

\begin{table}
	\centering
	\caption{Model Component Analysis.}
	\label{compo-eval}
	\resizebox{0.4\textwidth}{!}{
		\begin{tabular}{|l|c|c|}
			\hline
			& \multicolumn{2}{c|}{SNR(dB)}\\ \hline
			&Motion &ECG \\ \hline
			Initial Noise &12.57	 &-6.72               \\ \hline
			Encoder-Decoder (2+2 layers) & 21.06&3.36  \\ \hline
			\quad  + 3+3 layers           & 22.60 &3.49  \\ \hline
			\quad  \quad + dilated conv. / deconv.      & 24.47 &5.09\\ \hline
			\quad  \quad \quad + residual learning        & 25.21&4.24 \\ \hline
			\quad  \quad \quad \quad + adversarial learning     &\textbf{32.08}  &\textbf{5.30}\\ \hline
		\end{tabular}
	}
\end{table}

\section{Conclusion}\label{sec:conclusion}
We have presented signal denoising as a distribution alignment task based on adversarial learning. Our denoising tool is encoder-decoder deep neural network that processes signals represented by a sequence of measurements. In the evaluations, we show that our approach generalizes to different signal and noise types. Furthermore, we demonstrate better performance than learning-based methods and filtering approaches. As future work, we aim to study more sequential signals related to hand~\cite{nissler2015omg} and body pose estimation~\cite{belagiannis2014holistic}.

\bibliographystyle{IEEEtran}
\bibliography{egbib}

\begin{thebibliography}{10}
\providecommand{\url}[1]{#1}
\csname url@samestyle\endcsname
\providecommand{\newblock}{\relax}
\providecommand{\bibinfo}[2]{#2}
\providecommand{\BIBentrySTDinterwordspacing}{\spaceskip=0pt\relax}
\providecommand{\BIBentryALTinterwordstretchfactor}{4}
\providecommand{\BIBentryALTinterwordspacing}{\spaceskip=\fontdimen2\font plus
\BIBentryALTinterwordstretchfactor\fontdimen3\font minus
  \fontdimen4\font\relax}
\providecommand{\BIBforeignlanguage}[2]{{%
\expandafter\ifx\csname l@#1\endcsname\relax
\typeout{** WARNING: IEEEtran.bst: No hyphenation pattern has been}%
\typeout{** loaded for the language `#1'. Using the pattern for}%
\typeout{** the default language instead.}%
\else
\language=\csname l@#1\endcsname
\fi
#2}}
\providecommand{\BIBdecl}{\relax}
\BIBdecl

\bibitem{arsene2019deep}
C.~T. Arsene, R.~Hankins, and H.~Yin, ``Deep learning models for denoising ecg
  signals,'' in \emph{2019 27th European Signal Processing Conference
  (EUSIPCO)}.\hskip 1em plus 0.5em minus 0.4em\relax IEEE, 2019, pp. 1--5.

\bibitem{im2017denoising}
D.~I.~J. Im, S.~Ahn, R.~Memisevic, and Y.~Bengio, ``Denoising criterion for
  variational auto-encoding framework,'' in \emph{Thirty-First AAAI Conference
  on Artificial Intelligence}, 2017.

\bibitem{2010_vincent}
\BIBentryALTinterwordspacing
P.~Vincent, H.~Larochelle, I.~Lajoie, Y.~Bengio, and P.-A. Manzagol, ``Stacked
  denoising autoencoders: Learning useful representations in a deep network
  with a local denoising criterion,'' \emph{J. Mach. Learn. Res.}, vol.~11, pp.
  3371--3408, Dec. 2010. [Online]. Available:
  \url{http://dl.acm.org/citation.cfm?id=1756006.1953039}
\BIBentrySTDinterwordspacing

\bibitem{agostinelli2013adaptive}
F.~Agostinelli, M.~R. Anderson, and H.~Lee, ``Adaptive multi-column deep neural
  networks with application to robust image denoising,'' in \emph{NIPS}, 2013.

\bibitem{2016_mao}
X.~Mao, C.~Shen, and Y.-B. Yang, ``Image restoration using very deep
  convolutional encoder-decoder networks with symmetric skip connections,'' in
  \emph{NIPS}, 2016.

\bibitem{lu2013speech}
X.~Lu, Y.~Tsao, S.~Matsuda, and C.~Hori, ``Speech enhancement based on deep
  denoising autoencoder.'' in \emph{Interspeech}, 2013, pp. 436--440.

\bibitem{2016_vandenoord}
A.~van~den Oord, S.~Dieleman, H.~Zen, K.~Simonyan, O.~Vinyals, A.~Graves,
  N.~Kalchbrenner, A.~Senior, and K.~Kavukcuoglu, ``Wavenet: A generative model
  for raw audio,'' in \emph{Arxiv}, 2016.

\bibitem{grill2017two}
T.~Grill and J.~Schl{\"u}ter, ``Two convolutional neural networks for bird
  detection in audio signals,'' in \emph{2017 25th European Signal Processing
  Conference (EUSIPCO)}.\hskip 1em plus 0.5em minus 0.4em\relax IEEE, 2017, pp.
  1764--1768.

\bibitem{rethage2018wavenet}
D.~Rethage, J.~Pons, and X.~Serra, ``A wavenet for speech denoising,'' in
  \emph{2018 IEEE International Conference on Acoustics, Speech and Signal
  Processing (ICASSP)}.\hskip 1em plus 0.5em minus 0.4em\relax IEEE, 2018, pp.
  5069--5073.

\bibitem{antoniou2016digital}
A.~Antoniou, \emph{Digital signal processing}.\hskip 1em plus 0.5em minus
  0.4em\relax McGraw-Hill, 2016.

\bibitem{boudraa2004emd}
A.~Boudraa, J.~Cexus, and Z.~Saidi, ``Emd-based signal noise reduction,''
  \emph{Inter. Journal of Signal Processing}, vol.~1, no.~1, pp. 33--37, 2004.

\bibitem{kopsinis2009development}
Y.~Kopsinis and S.~McLaughlin, ``Development of emd-based denoising methods
  inspired by wavelet thresholding,'' \emph{IEEE Transactions on signal
  Processing}, vol.~57, no.~4, pp. 1351--1362, 2009.

\bibitem{huang1998empirical}
N.~E. Huang, Z.~Shen, S.~R. Long, M.~C. Wu, H.~H. Shih, Q.~Zheng, N.-C. Yen,
  C.~C. Tung, and H.~H. Liu, ``The empirical mode decomposition and the hilbert
  spectrum for nonlinear and non-stationary time series analysis,''
  \emph{Proceedings of the Royal Society of London. Series A: Mathematical,
  Physical and Engineering Sciences}, vol. 454, no. 1971, pp. 903--995, 1998.

\bibitem{huang2014hilbert}
N.~E. Huang, \emph{Hilbert-Huang transform and its applications}.\hskip 1em
  plus 0.5em minus 0.4em\relax World Scientific, 2014, vol.~16.

\bibitem{maaten2008visualizing}
L.~v.~d. Maaten and G.~Hinton, ``Visualizing data using t-sne,'' \emph{Journal
  of machine learning research}, vol.~9, no. Nov, pp. 2579--2605, 2008.

\bibitem{goodfellow2014generative}
I.~Goodfellow, J.~Pouget-Abadie, M.~Mirza, B.~Xu, D.~Warde-Farley, S.~Ozair,
  A.~Courville, and Y.~Bengio, ``Generative adversarial nets,'' in \emph{NIPS},
  2014.

\bibitem{makhzani2015adversarial}
A.~Makhzani, J.~Shlens, N.~Jaitly, I.~Goodfellow, and B.~Frey, ``Adversarial
  autoencoders,'' \emph{arXiv preprint arXiv:1511.05644}, 2015.

\bibitem{creswell2018denoising}
A.~Creswell and A.~A. Bharath, ``Denoising adversarial autoencoders,''
  \emph{IEEE transactions on neural networks and learning systems}, no.~99, pp.
  1--17, 2018.

\bibitem{long2015fully}
J.~Long, E.~Shelhamer, and T.~Darrell, ``Fully convolutional networks for
  semantic segmentation,'' in \emph{CVPR}, 2015.

\bibitem{chen2018deeplab}
L.-C. Chen, G.~Papandreou, I.~Kokkinos, K.~Murphy, and A.~L. Yuille, ``Deeplab:
  Semantic image segmentation with deep convolutional nets, atrous convolution,
  and fully connected crfs,'' \emph{IEEE transactions on pattern analysis and
  machine intelligence}, vol.~40, no.~4, pp. 834--848, 2018.

\bibitem{2018_bai}
\BIBentryALTinterwordspacing
S.~Bai, J.~Zico~Kolter, and V.~Koltun, ``{An Empirical Evaluation of Generic
  Convolutional and Recurrent Networks for Sequence Modeling},'' \emph{ArXiv
  e-prints}, 2018. [Online]. Available:
  \url{http://adsabs.harvard.edu/abs/2018arXiv180301271B}
\BIBentrySTDinterwordspacing

\bibitem{2016_he}
K.~He, X.~Zhang, S.~Ren, and J.~Sun, ``Deep residual learning for image
  recognition,'' \emph{CVPR}, 2016.

\bibitem{belagiannis2018adversarial}
V.~Belagiannis, A.~Farshad, and F.~Galasso, ``Adversarial network
  compression,'' in \emph{Computer Vision -- ECCV 2018 Workshops}.\hskip 1em
  plus 0.5em minus 0.4em\relax Springer International Publishing, 2019, pp.
  431--449.

\bibitem{tzeng2017adversarial}
E.~Tzeng, J.~Hoffman, K.~Saenko, and T.~Darrell, ``Adversarial discriminative
  domain adaptation,'' in \emph{CVPR}, 2017.

\bibitem{tagkey2009iii}
S.~Mallat, \emph{A wavelet tour of signal processing}.\hskip 1em plus 0.5em
  minus 0.4em\relax Elsevier, 1999.

\bibitem{greff2017lstm}
K.~Greff, R.~K. Srivastava, J.~Koutn{\'\i}k, B.~R. Steunebrink, and
  J.~Schmidhuber, ``Lstm: A search space odyssey,'' \emph{IEEE transactions on
  neural networks and learning systems}, vol.~28, no.~10, pp. 2222--2232, 2017.

\bibitem{2012_zeiler}
M.~D. Zeiler, ``Adadelta: An adaptive learning rate method,'' \emph{CoRR},
  2012.

\bibitem{2010_glorot}
X.~Glorot and Y.~Bengio, ``Understanding the difficulty of training deep
  feedforward neural networks,'' in \emph{AISTATS}, 2010.

\bibitem{2016_burri}
M.~Burri, J.~Nikolic, P.~Gohl, T.~Schneider, J.~Rehder, S.~Omari, M.~W.
  Achtelik, and R.~Siegwart, ``The euroc micro aerial vehicle datasets,''
  \emph{The International Journal of Robotics Research}, 2016.

\bibitem{goldberger2000physiobank}
A.~L. Goldberger, L.~A. Amaral, L.~Glass, J.~M. Hausdorff, P.~C. Ivanov, R.~G.
  Mark, J.~E. Mietus, G.~B. Moody, C.-K. Peng, and H.~E. Stanley, ``Physiobank,
  physiotoolkit, and physionet: components of a new research resource for
  complex physiologic signals,'' \emph{Circulation}, vol. 101, no.~23, pp.
  e215--e220, 2000.

\bibitem{nissler2015omg}
C.~Nissler, N.~Mouriki, C.~Castellini, V.~Belagiannis, and N.~Navab, ``Omg:
  introducing optical myography as a new human machine interface for hand
  amputees,'' in \emph{2015 IEEE International Conference on Rehabilitation
  Robotics (ICORR)}.\hskip 1em plus 0.5em minus 0.4em\relax IEEE, 2015, pp.
  937--942.

\bibitem{belagiannis2014holistic}
V.~Belagiannis, C.~Amann, N.~Navab, and S.~Ilic, ``Holistic human pose
  estimation with regression forests,'' in \emph{International Conference on
  Articulated Motion and Deformable Objects}.\hskip 1em plus 0.5em minus
  0.4em\relax Springer, 2014, pp. 20--30.

\end{thebibliography}

\end{document}